\documentclass{article}
\usepackage[utf8]{inputenc}
\usepackage{graphicx}
\usepackage{multirow}
\usepackage{url}
\usepackage{hyperref}
\usepackage{xcolor}
\usepackage{arxiv}

\title{ASAP: Adaptive Scheme for Asynchronous Processing of Event-based Vision Algorithms}

\author{
    \href{https://orcid.org/0000-0002-4435-5466}{\includegraphics[scale=0.06]{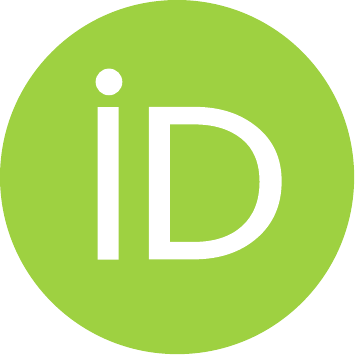}\hspace{1mm}Raul Tapia}\\
	GRVC Robotics Lab.\\
	Universidad de Sevilla\\
	\texttt{raultapia@us.es}\\
	\And
	\href{https://orcid.org/0000-0002-2285-2605}{\includegraphics[scale=0.06]{orcid.pdf}\hspace{1mm}Augusto Gómez Eguíluz} \\
	GRVC Robotics Lab.\\
	Universidad de Sevilla\\
	\texttt{ageguiluz@us.es}\\
	\And
	\href{https://orcid.org/0000-0001-9431-7831}{\includegraphics[scale=0.06]{orcid.pdf}\hspace{1mm}José Ramiro Martínez-de Dios} \\
	GRVC Robotics Lab.\\
	Universidad de Sevilla\\
	\texttt{jdedios@us.es}\\
	\And
	\href{https://orcid.org/0000-0003-2155-2472}{\includegraphics[scale=0.06]{orcid.pdf}\hspace{1mm}Anibal Ollero} \\
	GRVC Robotics Lab.\\
	Universidad de Sevilla\\
	\texttt{aollero@us.es}\\
}

\paperurl{https://raultapia.com/preprints/0001/redirect}
\journal{International Conference on Robotics and Automation 2020 Workshop on Unconventional Sensors in Robotics}
\editorial{IEEE}

\begin{document}

\maketitle

\begin{abstract}
Event cameras can capture pixel-level illumination changes with very high  temporal resolution and dynamic range. They have received increasing research interest due to their robustness to lighting conditions and motion blur. Two main approaches exist in the literature to feed the event-based processing algorithms: packaging the triggered events in \textit{event packages} and sending them one-by-one as \textit{single events}.
These approaches suffer limitations from either processing overflow or lack of responsivity. Processing overflow is caused by high event generation rates when the algorithm cannot process all the events in real-time. Conversely, lack of responsivity happens in cases of low event generation rates when the event packages are sent at too low frequencies.
This paper presents \textit{ASAP}, an adaptive scheme to manage the event stream through variable-size packages that accommodate to the event package processing times. The experimental results show that \textit{ASAP} is capable of feeding an asynchronous event-by-event clustering algorithm in a responsive and efficient manner and at the same time prevents overflow.
\end{abstract}

\section{Introduction}
\label{sec:intro}

The advent of event-based cameras has motivated increasing research interest in their application to robotics. 
Event cameras are neuromorphic sensors that capture the asynchronous illumination changes at pixel level and $\mu s$ resolution. They provide high dynamic range and are insensitive to motion blur. Additionally, they are lightweight and have low power consumption. A good number of successful techniques have been proposed in the last years evidencing their capabilities \cite{gallego2019event}.

The current trend in robotics is to group the received events in frames, i.e. \emph{event images}. Processing of \emph{event images} enables the design of complex and elaborated techniques, similar to those from traditional computer vision.
However, that approach does not always fully exploit the sequential and asynchronous capabilities of the data stream provided by event cameras. For instance, \emph{event images} can create motion blur in some cases and, some works, e.g. work \cite{Sanket2019EVDodgeEA} implemented specific mechanisms to mitigate it. Asynchronous \textit{event-by-event} processing usually has higher computational needs \cite{li2019fa, alzugaray2018asynchronous} and is sometimes combined with \emph{event images} for high-level processing, see e.g. \cite{vasco2017independent}.

Two main approaches exist in the literature to feed the event-based algorithms: using \textit{event packages} and \textit{single events}. Most of the existing frameworks, such as YARP \cite{Glover2017b} and Dynamic Vision System \cite{mueggler2014event}, rely on packaging the triggered events and transmitting them as output of the camera in packages. While these mechanisms prevent the algorithms' communication overhead, they are not efficient when the algorithm is capable of processing the events in one package before the next package is received. Differently, the work in \cite{marcireau2019sepia} presented a framework for \textit{single-event} handling that consists of three modules: an I/O library, a computation toolbox, and a visualization tool. Although event buffers were still used for camera communication, their framework suppresses the use of buffers between the tools.
Although \textit{single-event} delivery is efficient when the algorithm can process the events faster than they are received, feeding the algorithm at a temporal resolution of $\mu s$ can result in the computational overflowing.

\begin{figure}[ht]
\includegraphics[width=0.5\textwidth]{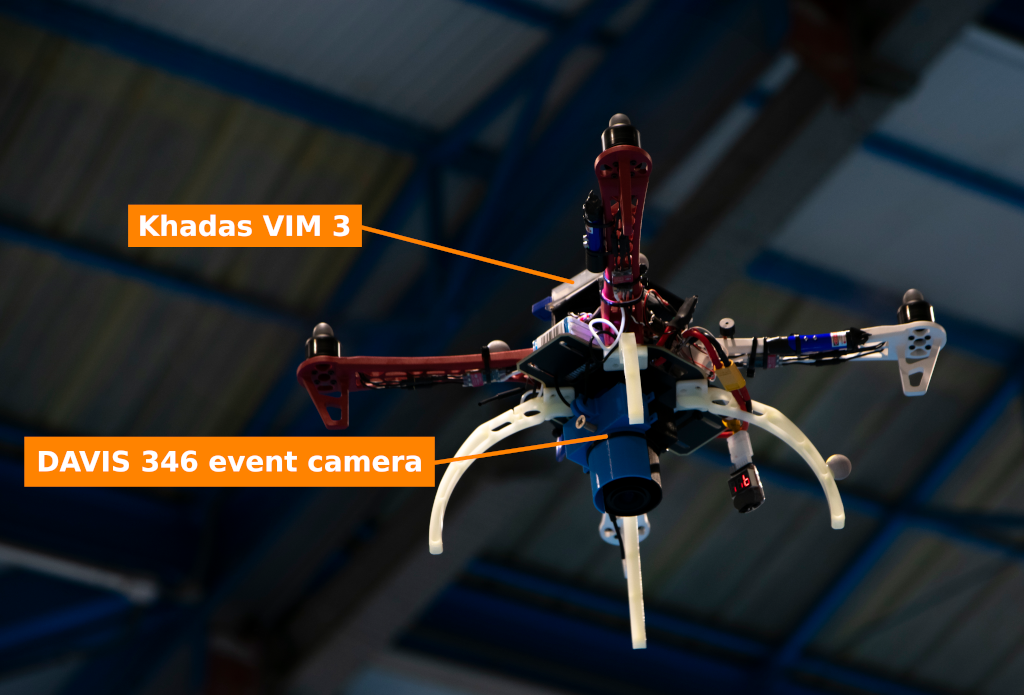}
\centering
\caption{Aerial robot based on \emph{DJI Flamewheel F450} equipped with a \emph{DAVIS 346} event camera and a low-cost \emph{Khadas VIM3} used for on-board computation.}
\label{fig:aerialPlatform}
\end{figure}

Therefore, there is a trade-off between responsiveness and the risk of system overflow when feeding the event-based algorithms using \textit{single events} or \textit{event packages}. 
While static scenes generate a low number of events per second and allow the use of some \textit{event-by-event} algorithms, dynamic and complex scenes often generate huge amounts of events in short periods. Additionally, event perception is largely influenced by the movement of the robot \cite{pequeno2019temporal}.
Therefore, real-time computation on-board a robot might not be possible for certain scenarios and hardware specifications, such as that used on the aerial robot shown in Fig.~\ref{fig:aerialPlatform}, which is the robot that has performed the experiments reported in this paper. In particular, the processing capability of aerial robots is limited mainly by their payload. In consequence, UAVs are often equipped with on-board computers with moderate computational resources, such as that installed on the aerial robot used in the reported experiments, which is shown in Fig.~\ref{fig:aerialPlatform}. This issue is even more relevant in flapping-wing robots since their payload limitations are more severe than in multi-rotors \cite{gomezeguiluz2019towards}.

This work presents \textit{ASAP}, a scheme to adapt the event packaging such that an asynchronous \textit{event-by-event} algorithm can process the events as soon as possible without overflowing.

\section{Method Description}
\label{sec:tracking}

The proposed method relies on two main mechanisms. First, the size of the \textit{event packages} is chosen according to the time required by the algorithm to process the last packages. Second, when the hardware limitations preclude the algorithm to process all events (i.e. for a maximum package size), some events are discarded in order to avoid overflowing.
Our work \cite{rodriguezgomez2020asynchronous} explored the effect of discarding events in a random manner to reduce computational cost and showed that the method could work using only $20\%$ of the full event stream without significantly affecting the algorithm performance. \textit{ASAP} adopts this approach and makes use of a random event discard procedure to reduce the risk of overflowing an even-by-event processing algorithm.

\begin{figure}[ht]
\includegraphics[width=0.75\textwidth]{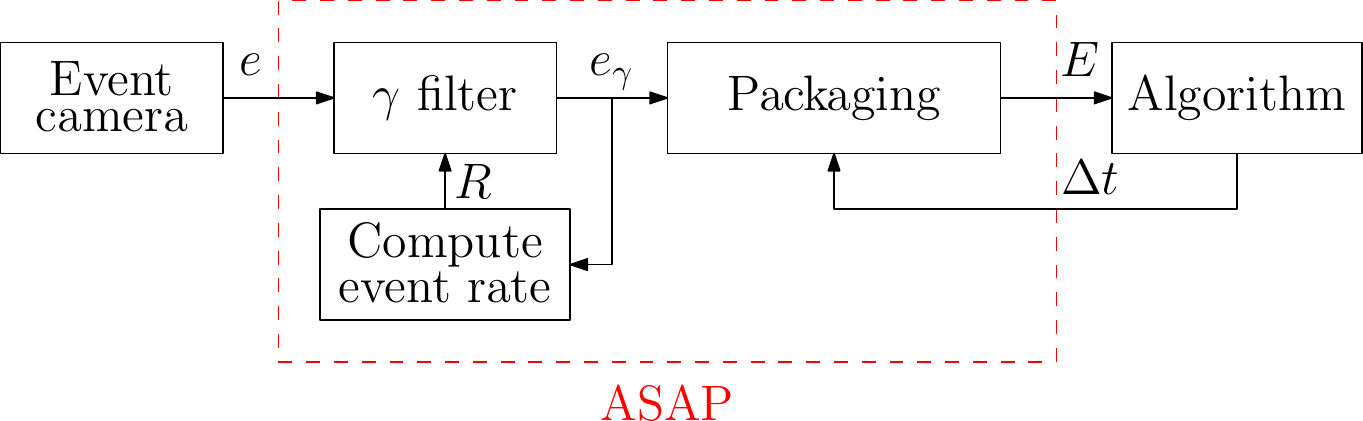}
\centering
\caption{Simplified scheme of \textit{ASAP} for adaptive event packaging.}
\label{fig:control_loop}
\end{figure}

Figure \ref{fig:control_loop} shows the proposed adaptive event packaging scheme. The first mechanism makes use of a close-loop approach to adapt the size of the event packages by considering time devoted by the asynchronous event-by-event algorithm to process the event package. A closed-loop mechanism established between modules \emph{Packaging} and \emph{Algorithm} adapts the size of the event packages according to the time required by the algorithm for processing the previous event package. The objective of this mechanism is to synchronize (i.e. make equal) the time required by the algorithm to process the events contained in the package and the temporal difference between the newest and oldest events in that package.

The second mechanism is designed for cases in which the rate of the triggered event stream is too high to be processed in real-time by the event-by-event processing algorithm. Relying on the findings in \cite{rodriguezgomez2020asynchronous}, this mechanism (i.e. \textit{$\gamma$ filter} module) performs a random event discard procedure to reduce the risk of algorithm overflowing. $R$ is the rate of events after event filtering by this module. The filtering rate $\gamma$ is dynamically adapted such that the filtered event rate $R$ is within a range that the algorithm can process in a responsive manner. If $R$ is higher than an upper bound $a$, the \textit{$\gamma$ filter} increases the number of discarded events using a simple adaptive criterion. $a$ was selected experimentally analyzing three types of scenarios with different event rates: low motion --low event rate-- moderate motion and high motion scenarios. $a$ was selected as an average event rate value suitable for most scenarios.

\section{Experimental Results}
\label{sec:experiments}

\begin{figure}[ht]
\includegraphics[height=0.2\textheight]{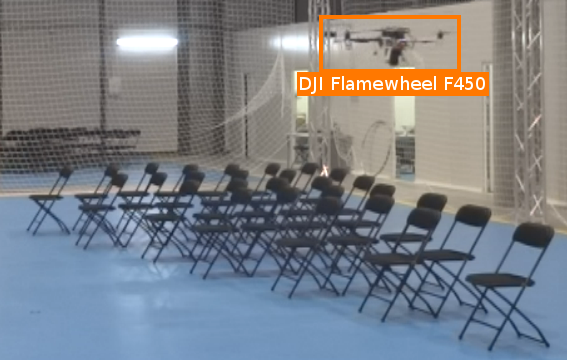}
\includegraphics[height=0.2\textheight]{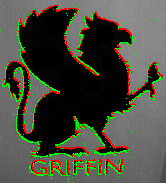}
\centering
\caption{Experimental scenarios. Left) Aerial robot used for the UAV on-board evaluation experiments. Right) Pattern used for the event overflow experiments.}
\label{fig:drone}
\end{figure}

\begin{figure}[ht]
\includegraphics[width=0.6\textwidth]{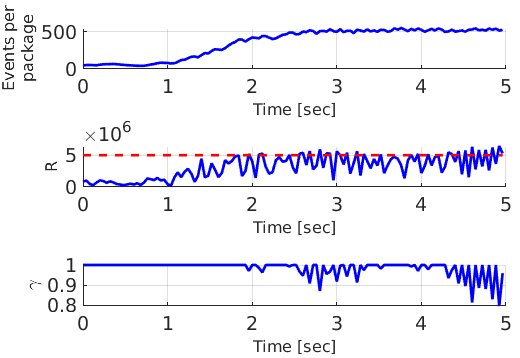}
\centering
\caption{Results in highly abrupt motion experiments.}
\label{fig:exp1}
\end{figure}

This section presents the experimental evaluation of \textit{ASAP} when processing the events with the asynchronous event-based clustering method presented in \cite{rodriguezgomez2020asynchronous}.
Two experiments were performed: highly abrupt camera motion and  UAV on-board evaluation. The first experiment was performed manually using a \emph{DAVIS346} event camera oriented such that the pattern in Fig. \ref{fig:drone}-right was in the camera field of view during the whole experiment. The vibration speed of the camera was increased progressively during $5$ s. Figure \ref{fig:exp1} shows the number of events per package, the value of $R$ and the value of $\gamma$ along the experiment. We selected $a = 5 \cdot 10^6$. With low vibration level, packages included very few events. As vibration level increased, more and more events were transmitted in each package using $\gamma=1$. When the event rate exceeded the boundary $a = 5 \cdot 10^6$, $\gamma$ was reduced to balance the filtered event rate ($R$) with the time required by the algorithm to process the event package running on the specific hardware, and thus adapting to the particular implementation.

\begin{figure}[ht]
\includegraphics[width=0.6\textwidth]{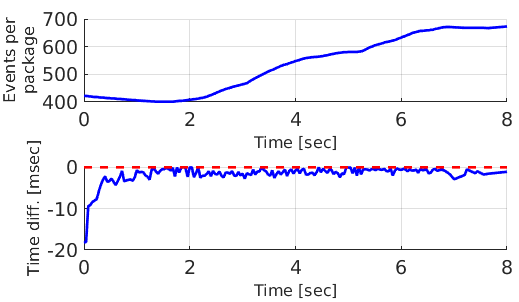}
\centering
\caption{Experimental results in UAV on-board evaluation.}
\label{fig:exp2}
\end{figure}

Additionally, the performance of \textit{ASAP} was evaluated on-board an UAV in an indoor scenario as shown in Figure \ref{fig:drone}-left. The experimental setup consists of a \emph{DAVIS346} event camera installed on-board a \textit{DJI Flamewheel F450} frame with a \textit{PixRacer} autopilot (see Fig. \ref{fig:aerialPlatform}). A low-cost \textit{Khadas VIM3} board is used for real-time running the event-based clustering method described in \cite{rodriguezgomez2020asynchronous} and for logging the results. \textit{ASAP} was implemented in C++ on top of the UAL abstraction layer \cite{real2018ual} using \textit{ROS Kinetic} and the \textit{PX4} low-level controller. In the experiment, the UAV flew over several chairs as shown in Fig. \ref{fig:drone}-left. The number of chairs increased as the UAV moved towards the goal and hence, the number of events generated over time also increased (see Fig. \ref{fig:exp2}-top). However, the event rate never exceeded the threshold $a$ and, therefore, the full event stream (i.e. $\gamma=1$, no events were discarded) was packaged --and also on-line processed by the  clustering algorithm-- during the whole experiment. 

Moreover, Fig. \ref{fig:exp2}-bottom shows the difference between the time required by the clustering algorithm described in  \cite{rodriguezgomez2020asynchronous} to process a package and the time difference between the newest and oldest events in that package. As long as the time difference
remains negative, real-time processing is guaranteed as the algorithm will always process a package before the next one is received. Thus, \textit{ASAP} is capable of adapting the size of the packages to keep the time difference negative --preventing the algorithm from overflowing-- and near zero --synchronizing algorithm executing and event packaging, which ensures responsive event processing.

\section{Conclusions and Future Work}
\label{sec:conclusion}

This paper presents \textit{ASAP}, an adaptive scheme for dynamic event packaging that enables responsive event processing while preventing processing overflow. The experimental results show that the proposed approach regulates the event stream fed to an \textit{event-by-event} clustering algorithm and it is capable of filtering upon saturation. Our future work will focus on enhancing the scheme response to abrupt changes in the event stream and on evaluating the scheme in a wider range of asynchronous event-based algorithms.

\section*{Acknowledgment}
This work was supported by the European Project GRIFFIN ERC Advanced Grant 2017, Action 788247. Partial funding was obtained from the Plan Estatal de Investigación Científica y Técnica y de Innovación of the Ministerio de Universidades del Gobierno de España (FPU19/04692).

\bibliographystyle{IEEEtran}
\bibliography{icra2020ws}

\begin{thebibliography}{10}
\providecommand{\url}[1]{#1}
\csname url@samestyle\endcsname
\providecommand{\newblock}{\relax}
\providecommand{\bibinfo}[2]{#2}
\providecommand{\BIBentrySTDinterwordspacing}{\spaceskip=0pt\relax}
\providecommand{\BIBentryALTinterwordstretchfactor}{4}
\providecommand{\BIBentryALTinterwordspacing}{\spaceskip=\fontdimen2\font plus
\BIBentryALTinterwordstretchfactor\fontdimen3\font minus
  \fontdimen4\font\relax}
\providecommand{\BIBforeignlanguage}[2]{{%
\expandafter\ifx\csname l@#1\endcsname\relax
\typeout{** WARNING: IEEEtran.bst: No hyphenation pattern has been}%
\typeout{** loaded for the language `#1'. Using the pattern for}%
\typeout{** the default language instead.}%
\else
\language=\csname l@#1\endcsname
\fi
#2}}
\providecommand{\BIBdecl}{\relax}
\BIBdecl

\bibitem{gallego2019event}
G.~Gallego, T.~Delbruck, G.~Orchard, C.~Bartolozzi, B.~Taba, A.~Censi,
  S.~Leutenegger, A.~Davison, J.~Conradt, K.~Daniilidis \emph{et~al.},
  ``Event-based vision: A survey,'' \emph{arXiv preprint arXiv:1904.08405},
  2019.

\bibitem{Sanket2019EVDodgeEA}
N.~J. Sanket, C.~M. Parameshwara, C.~D. Singh, A.~V. Kuruttukulam,
  C.~Ferm{\"u}ller, D.~Scaramuzza, and Y.~Aloimonos, ``Evdodge: Embodied ai for
  high-speed dodging on a quadrotor using event cameras,'' \emph{arXiv preprint
  arXiv:1906.02919}, 2019.

\bibitem{li2019fa}
R.~Li, D.~Shi, Y.~Zhang, K.~Li, and R.~Li, ``Fa-harris: A fast and asynchronous
  corner detector for event cameras,'' in \emph{2019 IEEE/RSJ International
  Conference on Intelligent Robots and Systems (IROS)}.\hskip 1em plus 0.5em
  minus 0.4em\relax IEEE, 2019.

\bibitem{alzugaray2018asynchronous}
I.~Alzugaray and M.~Chli, ``Asynchronous corner detection and tracking for
  event cameras in real time,'' \emph{IEEE Robotics and Automation Letters},
  vol.~3, no.~4, pp. 3177--3184, 2018.

\bibitem{vasco2017independent}
V.~Vasco, A.~Glover, E.~Mueggler, D.~Scaramuzza, L.~Natale, and C.~Bartolozzi,
  ``Independent motion detection with event-driven cameras,'' in \emph{2017
  18th International Conference on Advanced Robotics (ICAR)}.\hskip 1em plus
  0.5em minus 0.4em\relax IEEE, 2017, pp. 530--536.

\bibitem{Glover2017b}
A.~Glover, V.~Vasco, M.~Iacono, and C.~Bartolozzi, ``{The event-driven Software
  Library for YARP — With Algorithms and iCub Applications},''
  \emph{Frontiers in Robotics and AI}, vol.~4, p.~73, 2018.

\bibitem{mueggler2014event}
E.~Mueggler, B.~Huber, and D.~Scaramuzza, ``Event-based, 6-dof pose tracking
  for high-speed maneuvers,'' in \emph{2014 IEEE/RSJ International Conference
  on Intelligent Robots and Systems}, 2014, pp. 2761--2768.

\bibitem{marcireau2019sepia}
A.~Marcireau, S.~H. Ieng, and R.~B. Benosman, ``Sepia, tarsier and chameleon: a
  modular c++ framework for event-based computer vision.'' \emph{Frontiers in
  Neuroscience}, vol.~13, p. 1338, 2019.

\bibitem{pequeno2019temporal}
A.~Peque{\~n}o-Zurro, D.~Shaikh, and I.~Ra{\~n}{\'o}, ``Temporal changes in
  stimulus perception improve bio-inspired source seeking,'' \emph{arXiv
  preprint arXiv:1903.10279}, 2019.

\bibitem{gomezeguiluz2019towards}
A.~G\'{o}mez~Egu\'{i}luz, J.~P. Rodr\'{i}guez-G\'{o}mez, J.~Paneque, P.~Grau,
  J.~R. Martinez De-Dios, and A.~Ollero, ``Towards flapping wing robot visual
  perception: Opportunities and challenges,'' in \emph{IEEE RED-UAS}, 2019.

\bibitem{rodriguezgomez2020asynchronous}
J.~P. Rodr\'{\i}guez-G\'omez, A.~G\'omez~Egu\'{\i}luz, J.~R. Martínez-de Dios,
  and A.~Ollero, ``Asynchronous event-based clustering and tracking for
  intrusion monitoring in uas,'' in \emph{IEEE International Conference on
  Robotics and Automation (ICRA)}.\hskip 1em plus 0.5em minus 0.4em\relax IEEE,
  2020.

\bibitem{real2018ual}
F.~Real, A.~Torres-Gonz{\'a}lez, P.~Ram{\'o}n-Soria, J.~Capit{\'a}n, and
  A.~Ollero, ``Ual: An abstraction layer for unmanned aerial vehicles,'' in
  \emph{2nd Intl. Symposium on Aerial Robotics}, 2018.

\end{thebibliography}

\end{document}